\documentclass[]{llncs}

\usepackage[T1]{fontenc}
\usepackage{graphicx}
\usepackage{listings}
\usepackage{subfig}
\usepackage{tabularx}

\begin{document}
\title{INPUT 2022\\ Team Description Paper}
%
%
\author{Masaki Yasuhara \and Tomoya Takahashi 
\and Hiroki Maruta \and Hiroyuki Saito \and Shota Higuchi 
\and Takaaki Nara \and Keitaro Takeuchi \and Yota Sakai \and Kazuki Ishibashi}
%
%
\institute{Nagaoka Activation Zone of Energy, 4-1-9, Shinsan Nagaoka, Niigata, 9402127 JAPAN\\
\email{Email: rcjinput@gmail.com}\\
\url{URL: https://input-ssl.dev/}}
\maketitle              

\begin{abstract}
INPUT is a team participating in the RoboCup Soccer Small League (SSL). 
It aims to show the world the technological capabilities of the Nagaoka region of Niigata Prefecture, which is where the team members are from. 
For this purpose, we are working on one of the projects from the Nagaoka Activation Zone of Energy (NAZE). 
Herein, we introduce two robots, v2019 and v2022, as well as AI systems that will be used in RoboCup 2022.
In addition, we describe our efforts to develop robots in collaboration with companies in the Nagaoka area. 
\end{abstract}

\section{Introduction}
\begin{figure}[htbp]
    \centering
    \begin{tabular}{c}
        \begin{minipage}{0.48\linewidth}
            \centering
            \includegraphics[width=0.95\linewidth]{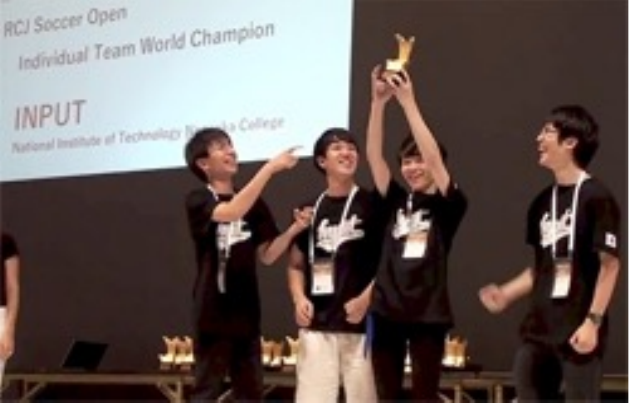}
            \caption{World Championship in the Junior League.}
            \label{fig:junior}
        \end{minipage}
        \hfill
        \begin{minipage}{0.48\linewidth}
            \centering
            \includegraphics[width=0.95\linewidth]{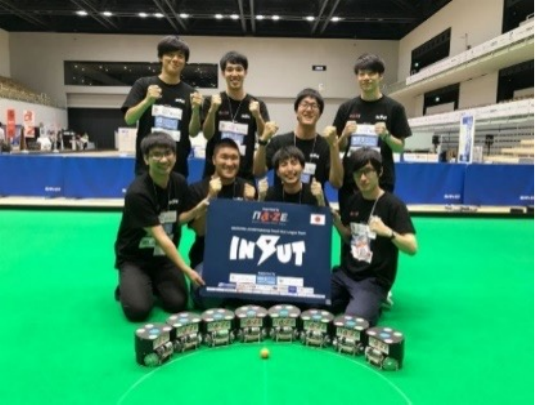}
            \caption{RoboCup Japan Open 2019, Nagaoka.}
            \label{fig:jo2019}
        \end{minipage}
    \end{tabular}
\end{figure}

Team INPUT was formed in August 2018.
The team is unique in the aspect that many of its members have prior experience of participation in the RoboCup Junior Soccer Open.
Our team consists of some members from three junior teams: CatPot, INPUT, and CatBot. These teams have participated in world competitions, and INPUT in particular won the RoboCup 2017 Nagoya (Fig. \ref{fig:junior}).
All members are graduates of the National Institute of Technology, Nagaoka College (NITNC).
INPUT is one of NAZE's projects from the Nagaoka area, where NAZE and NITNC are located. Nagaoka is a manufacturing town.
We are developing robots in collaboration with companies in the Nagaoka area.
The RoboCup 2019 Japan Tournament held in Nagaoka City was INPUT's first official competition (Fig. \ref{fig:jo2019}).
In the first game of the tournament, all eight robots worked and successfully passed the ball. We were able to win one game.

\section{Introducing Robot Generation v2019}
\begin{figure}[ttbp]
    \centering
    \includegraphics[width=0.9\linewidth]{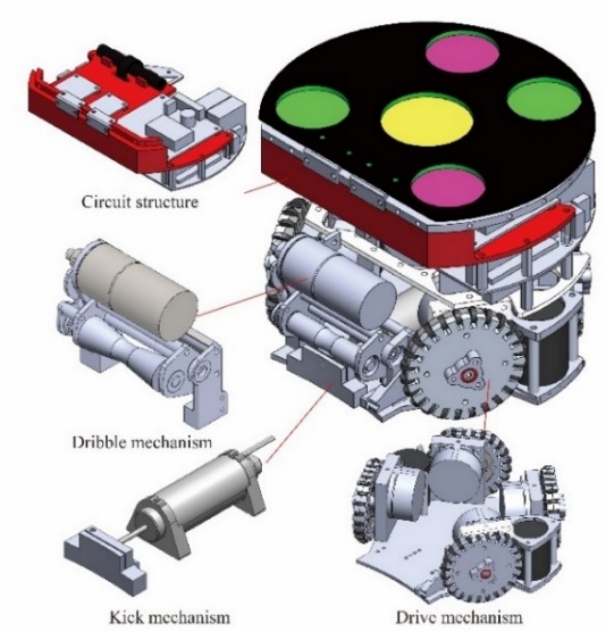}
    \caption{Components of robot v2019.}
    \label{fig:robot2019}
\end{figure}

This chapter describes the configuration of the v2019 robot used in RoboCup 2019 Japan Open, Nagaoka.
This robot was the first robot developed by team INPUT.
Fig. \ref{fig:robot2019} shows the overall view of the v2019 robot.
\subsection{Mechanism Design}
\begin{figure}[ttbp]
    \centering
    \includegraphics[width=0.5\linewidth]{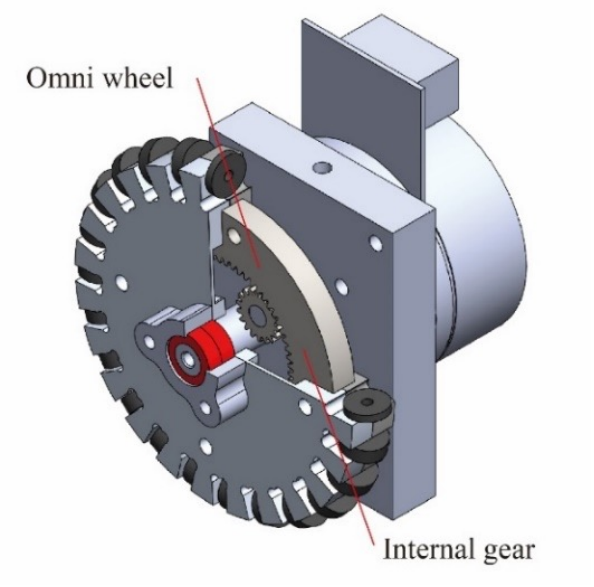}
    \caption{Drive unit.}
    \label{fig:dribeunit2019}
\end{figure}

The drive system of this robot is composed of four motor units (Fig. \ref{fig:dribeunit2019}).
These mechanical units were designed with consideration for replaceability and ease of repair, such that the units can be easily removed by unscrewing three bolts.
In this drive mechanism, the rotation is decelerated and transmitted by the internal gear attached to the omni-wheel.
By using the internal gear, it is possible to reduce the distance between the motor and the rotation axis of the wheel, and thus possible to lower the mounting position of the motor and lower the center of gravity of the robot.
The reduction ratio of the internal gear was approximately 3.33: 1, and the maximum tire speed was 1557 rpm.

\begin{figure}[htbp]
    \centering
    \begin{tabular}{c}
        \begin{minipage}{0.48\linewidth}
            \centering
            \includegraphics[width=0.95\linewidth]{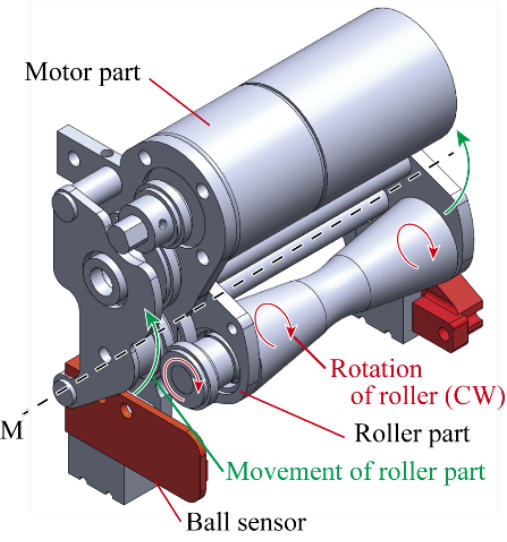}
            \caption{Dribble mechanism.}
            \label{fig:dribbler2019}
        \end{minipage}
        \hfill
        \begin{minipage}{0.48\linewidth}
            \centering
            \includegraphics[width=0.95\linewidth]{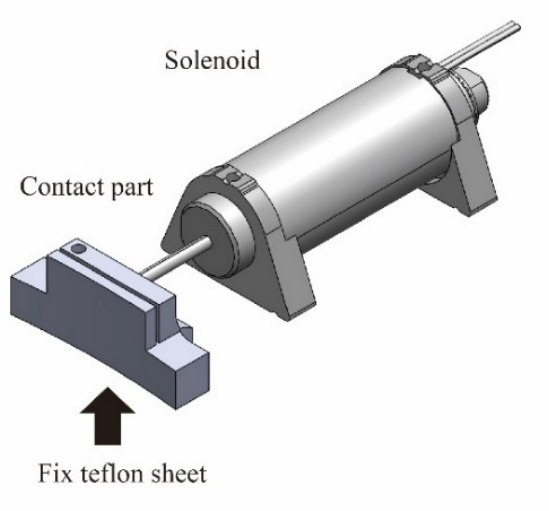}
            \caption{Kick mechanism}
            \label{fig:kicker2019}
        \end{minipage}
    \end{tabular}
\end{figure}

This robot is equipped with a deformable dribble mechanism, as shown in Fig. \ref{fig:dribbler2019}.
The dribble unit is composed of two parts: the motor and the roller.
The roller part is movable around the rotation axis M. This is because when the ball comes into contact with the roller, the impact is released by the movement of the roller.

For the kick mechanism of this robot, we selected the Super Stroke Solenoid (S-1012SS, Shindengen Mechatronics Co., Ltd.), which has a sufficient output and a long stroke.
A Teflon sheet was attached to the bottom of the contact part with the ball and was used as a guide (Fig. \ref{fig:kicker2019}).

%
%

\subsection{Circuit}
\begin{figure}[ttbp]
    \centering
    \includegraphics[width=0.9\linewidth]{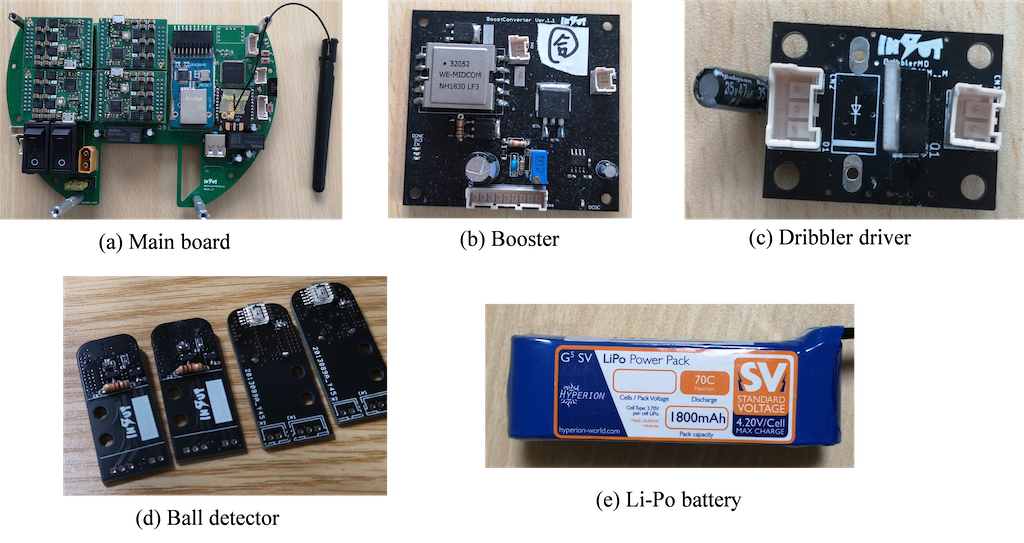}
    \caption{Circuit components.}
    \label{fig:circuit2019}
\end{figure}

The circuit of this robot consists of five parts: the main board, voltage boost circuit, motor driver for the dribbler, ball detector, and the Li-Po battery.
The main board contains the motor driver for the omni-wheel and communication modules.
The details are shown in Fig. \ref{fig:circuit2019} and Table \ref{tab:circuit2019}.
\begin{table}[htbp]
\centering
\caption{List of circuit components and their details}
\label{tab:circuit2019}
\begin{tabularx}{\linewidth}{X|X}
\hline
Device & Description \\\hline
Main board (Fig. \ref{fig:circuit2019}a) & CPU is STM32F407VGT6 on 215 MHz. \\
 & It contains 4 motor drivers ESCON50/5. \\
 & An XBee3 is used for communication with the PC. \\
 & DP83848 ethernet phi-module is included for using Wi-Fi router. \\
 \hline
 Booster (Fig. \ref{fig:circuit2019}b) & Li-Po battery 4 cell is boosted to 175 V to use the kicker. \\
\hline
Motor driver for dribbler (Fig. \ref{fig:circuit2019}c) & An Nch MOSFET (EKI04027) DC motor driver. \\
\hline
Ball detector (Fig. \ref{fig:circuit2019}d) & Photo IC with optical switch functions (S8119) and Infrared LED. \\
\hline
5 GHz WIFI router & WRH-583BK2-S, IEEE 802.11abgn 2.4/5 GHz wireless LAN \\
\hline
Li-Po battery (Fig. \ref{fig:circuit2019}e) & Hyperion LiPo 4-cell 1800 mAh \\
\hline
\end{tabularx}
\end{table}

The STM32 was developed using the hardware abstraction layer (HAL) library.
An XBee 2.4 GHz band radio was used to communicate with the PC.
When the XBee was unavailable, 5 GHz Wi-Fi was used to communicate with the PC.
The moving motor does not have an external encoder, and the ESCON is voltage-controlled; the CPU conveys the speed of the motor to the ESCON through pulse width modulation (PWM) signals.
The lack of an external encoder is due to space constraints inside the robot.
The motor of the dribbler uses a single FET and can only rotate in the direction of the ball roll.
The dribbler does not need to rotate in the counterclockwise (CCW) direction; hence, a very simple circuit is used to rotate only in the clockwise (CW) direction.

The voltage boost circuit of this robot is based on the voltage boost circuit of Team Roots \cite{roots}, and it is boosted up to 175 V using LT3750. The kicking force is adjusted based on the switching speed of the IGBT.

The robot is equipped with a ball sensor in the dribbling mechanism, as shown in red in Fig. \ref{fig:dribbler2019}, which is used in order to determine if the ball was ready to be kicked.
A ball sensor was mounted on the dribble mechanism.
A photo IC (S8119) and an infrared LED were used to detect the presence of the ball in the dribbler mechanism.

\begin{figure}[ttbp]
    \centering
    \includegraphics[width=0.9\linewidth]{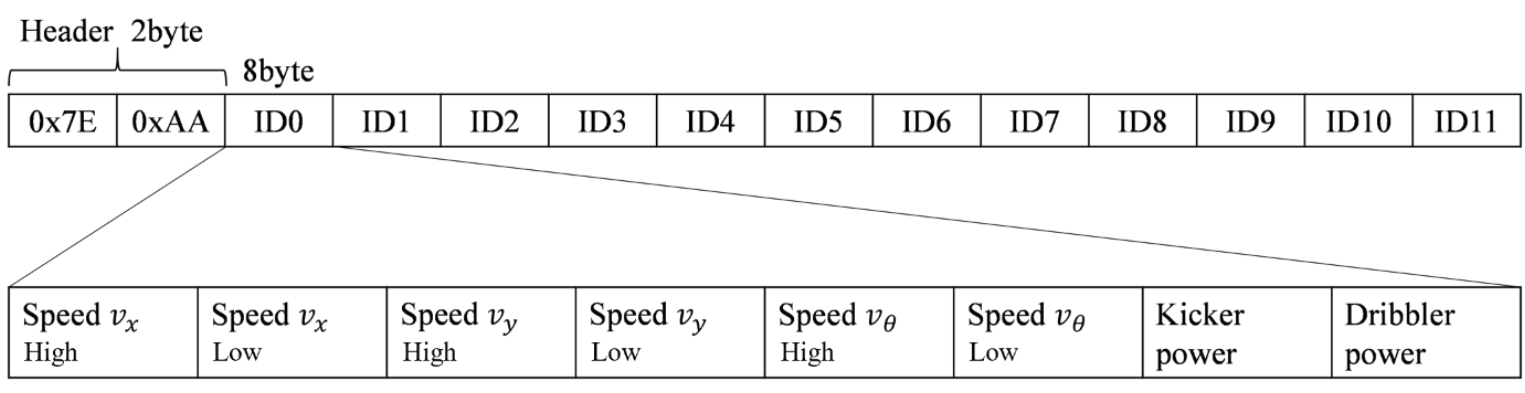}
    \caption{Communication protocol for XBee.}
    \label{fig:protocol2019}
\end{figure}

The communication protocol for XBee is shown in Fig. \ref{fig:protocol2019}. XBee operates in the AT mode with a baud rate of 115200 bps.
The speeds $v_x,v_y,v_{\theta}$ are received as floats.
The 5 GHz Wi-Fi is used to receive the User Datagram Protocol (UDP) using FreeRTOS and LwIP.
The Wi-Fi router assigns a local IP address to each robot and sends a comma separated string to the specified IP address (each robot is assigned a local IP address by the router, and a comma-separated string is sent to the specified IP address (robot)).
The contents of the packet are $v_x,v_y,v_{\theta}$, kicker power, and dribbler power.
%
%

\section{Introducing Robot Generation v2022}
While the v2019 robot had sufficient functionality to work in a match, it had a high center of mass (COM) that caused the robot to nearly fall over when accelerating from a stationary state.
We developed v2022, which solves this problem by improving the hardware.
In the following section, we explain the solutions for each problem.
\subsection{Improved Wheels for a Lower COM of Robot}
\begin{figure}[htbp]
    \centering
    \subfloat[][]{\includegraphics[width=0.45\linewidth]{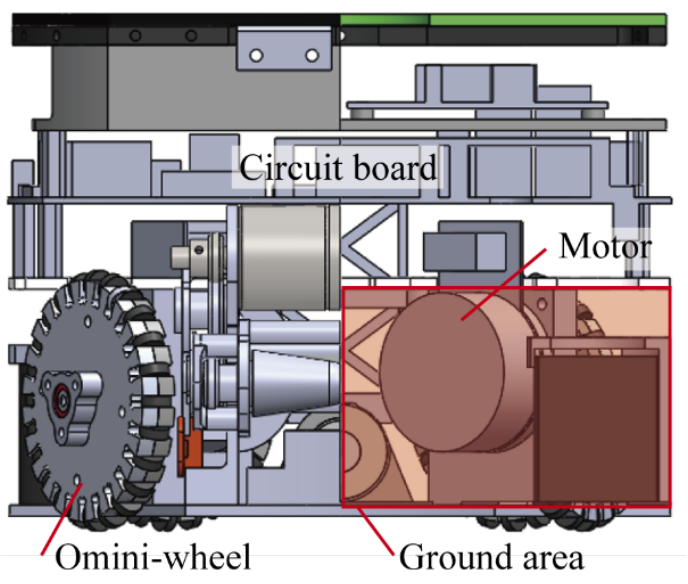}}
    \subfloat[][]{\includegraphics[width=0.45\linewidth]{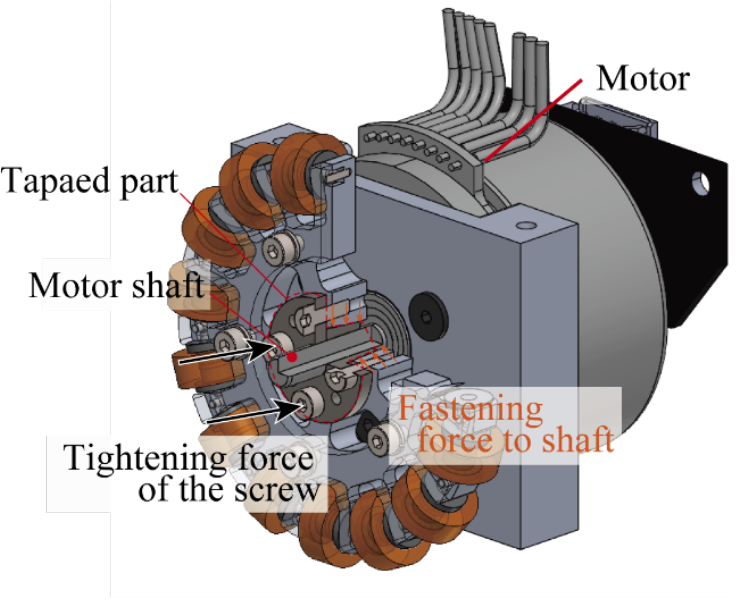}} \quad
    \subfloat[][]{\includegraphics[width=0.45\linewidth]{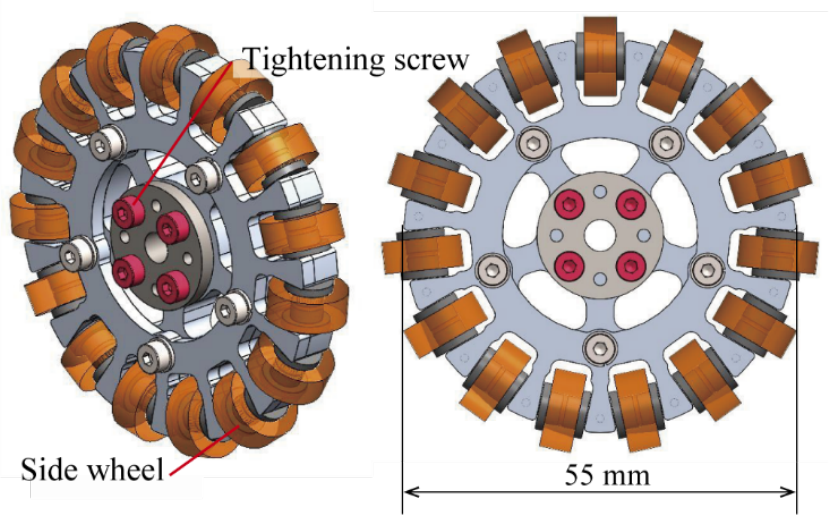}}
    \subfloat[][]{\includegraphics[width=0.45\linewidth]{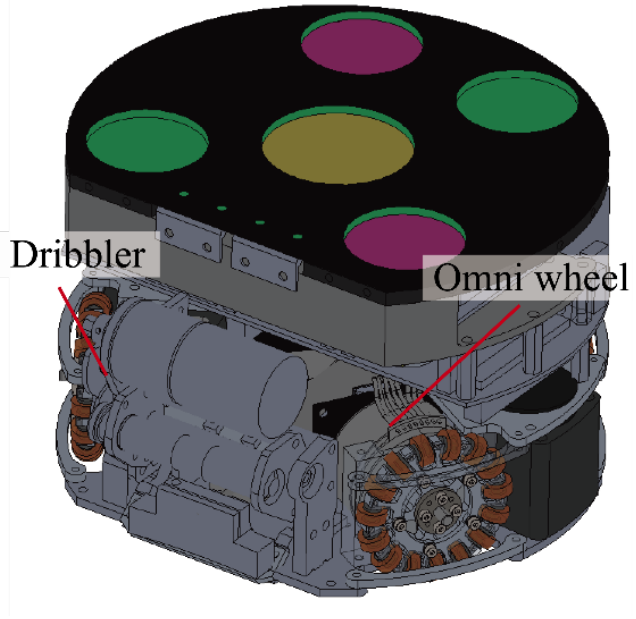}}
    \caption{v2022 robot with the new Omni-Wheel. (a) Ground area of robot, (b) Cross-section view of drive unit, (c) Wheel unit, (d) Isometric view of v2022.}
    \label{fig:robot2021}
\end{figure}

To lower the COM, we focused on enlarging the space closer to the ground in the configuration of the robot (ground area), as shown in Fig. \ref{fig:robot2021}(a).
Heavy objects must be placed at a lower height to lower the COM, and sufficient space in the lower area is needed to achieve this.
However, because this area is at a height where the ball can easily be touched, devices that directly touch the ball (solenoids for kicking, sensors, etc.) as well as the motor wheels for driving are placed here.
In some teams, the wheel drive motor is placed at the top for space reasons, and gears are used to transmit rotation to the wheels, resulting in a high COM.
Therefore, to reduce the COM, it is necessary to save space for these devices. Hence, we first made the omni-wheel thinner to increase the available space.

The bottleneck for making a thinner wheel lies in the process of fastening the wheel to the motor shaft.
Many teams inserted a screw from the side of the shaft and pressed down on the shaft to fix the wheel.
However, the thinner the wheel, the less fastening force is available because screws with a larger diameter cannot be used.
This method also requires access to the screws on the side of the wheel when attaching and detaching it, which increases the assembly time.
In Team-TIGERs \cite{tigers2020}, the shaft is fixed with glue, and the wheel is fixed with screws; however, the screws loosen due to vibration and rotation.

Therefore, we developed an omni-wheel equipped with a mechanical locking mechanism, as shown in Fig. \ref{fig:robot2021}(b).
The mechanical locking mechanism is a mechanical component that generates a force in the direction of diameter contraction by tightening a pair of tapered parts in the axial direction, thereby fixing the shaft and wheel by friction.
It can also be detached repeatedly by applying a force in the axial direction with the removal screw.
This enables the use of fewer parts, high fastening force, and simple attachment and removal. The basic characteristics of the developed omni-wheels are listed in Table 2.

In v2022, the wheel was directly connected to the motor shaft to increase the space in the lower section (Fig. \ref{fig:robot2021}(a)).
We used a Maxon motor EC 45 flat 70 W 24 V instead of the conventional Maxon motor EC 45 flat 50 W 18 V because it requires a large torque.
The circuit changes due to the change in the motor are described in the next section.
\begin{table}[htbp]
\centering
\caption{List of circuit components and their details}
\label{tab:v2019v2022}
\begin{tabularx}{\linewidth}{X|X|X}
\hline
 & v2019 & v2022 \\\hline
 Diameter [mm] & 80 & 55 \\
 Thickness [mm] & 16.5 & 11 \\
 Height of COM [mm] & 49.2 & 39.7 \\
 Material & A2017 & A2017 \\
\hline
\end{tabularx}
\end{table}

\subsection{Change of battery and voltage boost circuit}
The battery was changed to a 6-cell LiPo because the motor rating was changed to 24 V in v2022, as described in Section 4.1. 

\begin{figure}[htbp]
    \centering
    \subfloat[][]{\includegraphics[width=0.45\linewidth]{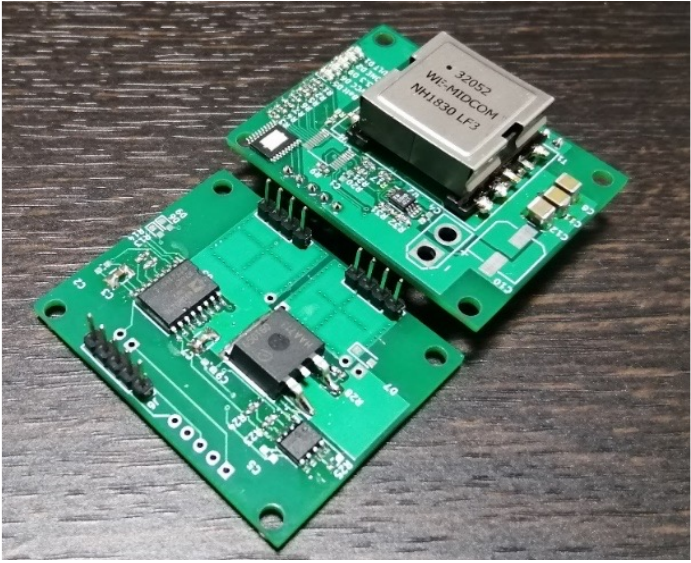}}
    \subfloat[][]{\includegraphics[width=0.45\linewidth]{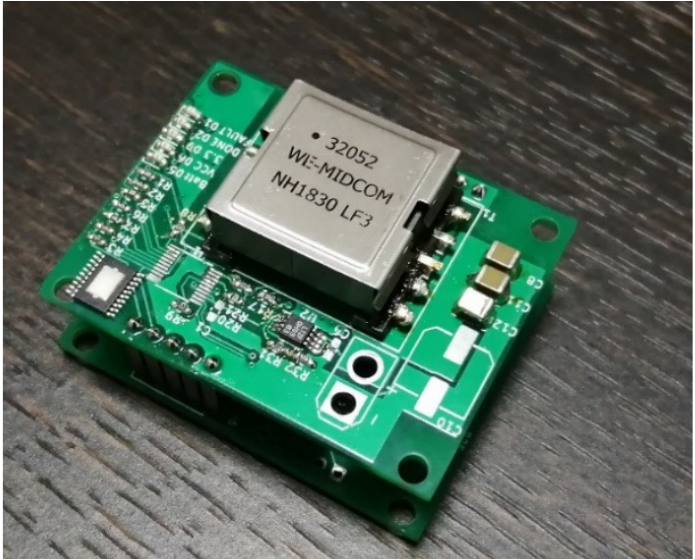}}
    \caption{v2022 Booster Circuit (a) separated, (b) stacked.}
    \label{fig:booster2021}
\end{figure}

The v2019 booster circuit is designed for a 4-cell LiPo battery and, hence, cannot be used in v2022.
In v2022, the voltage booster circuit is redesigned to allow a maximum input voltage of 40 V using LT3751 as the voltage booster control IC and an additional regulator.
The input circuit to the solenoid and the adjustment of the kick force were the same as in v2019. Fig. \ref{fig:booster2021} shows the v2022 boost circuit.
The occupied area was reduced using a two-stage board.

\section{Software}
\subsection{Structure}
\begin{figure}[ttbp]
    \centering
    \includegraphics[width=0.9\linewidth]{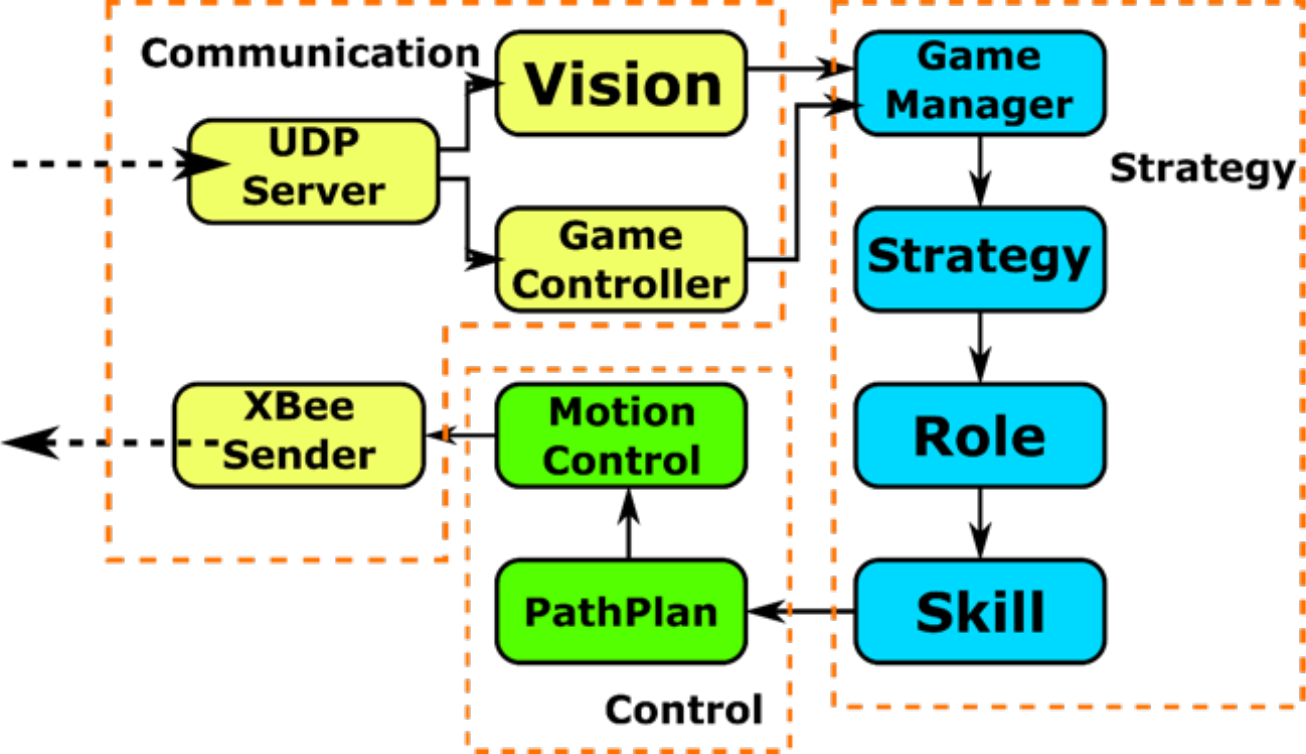}
    \caption{Software structure overview.}
    \label{fig:softwarestruct}
\end{figure}

The structure of the AI software is illustrated in Fig. \ref{fig:softwarestruct}.
It is divided into three major blocks: communication, strategy, and control.
The system is event-driven, triggered by the frame information in the ssl-vision. The entire sequence of operations, including the transmission from the trigger to the robot is processed within one frame.
First, the communication block receives the vision and referee information and sends a signal to the strategy block.
This includes communication with robots. The strategy block determines the movement of the robot based on the situation of the game.
The control block is responsible for path generation, acceleration/deceleration control, and yaw angle control to achieve the determined motion.

The strategy block consists of four modules. The GameManager module updates the status of the game from the vision and referee information.
The strategy module determines the combination of roles assigned to the robot and the priority of each role based on the situation of the match.
The role module appropriately assigns roles to robots and determines the actions to be taken for each role. The skill module is a subdivision of roles and consists of the most basic descriptions.

The control block consists of the PathPlan module and the MotionControl module.
The PathPlan module generates paths for collision avoidance. The MotionControl module performs acceleration/deceleration control and posture control.

The acceleration/deceleration control is based on the trapezoidal acceleration. The details are described in section 5.4. Yaw angle control was performed using PD control.

\subsection{Role}
One of the six roles is assigned to each robot. The number of roles assigned and their priorities depends on the situation of the match. 
\begin{description}
    \item[Goalie] This role is always assigned the highest priority, and a robot assigned this role only works as a goalie.
    \item[Attacker] This is a role that involves actively approaching the ball and aiming to pass, clear, and shoot; the attacker role is never assigned to more than one robot.
    \item[Defender] The robots assigned this role move around the perimeter of their own penalty area to block the path of shots. The number of defenders assigned to each team varies depending on the situation of the game. 
    \item[PassReceiver]  This is a role that involves waiting for a pass at a location where it is most likely to be received. The number of passengers assigned varies depending on the match situation. 
    \item[PassInterrupter] This player marks the opponent and intercepts the pass. The number of passengers assigned varies depending on the match situation. 
    \item[Waiter] This is the lowest priority role, and involves moving to a larger space for a smooth transition to another role.
 \end{description}

\subsection{Passing position}
The best location for passing is determined by the potential method, with the field delimited by a grid.

\begin{figure}[ttbp]
    \centering
    \includegraphics[width=0.9\linewidth]{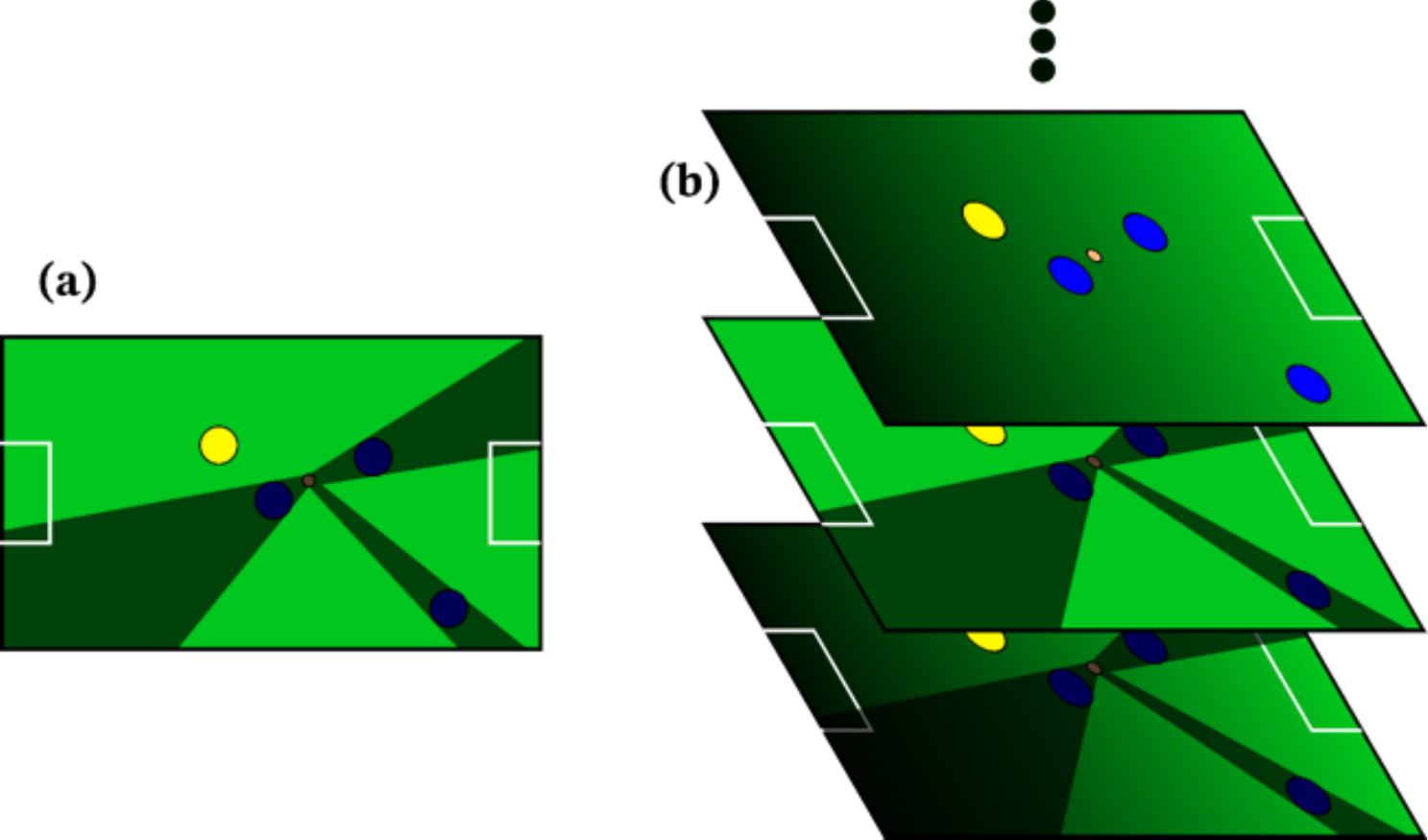}
    \caption{Potential score mask for determining the passing position. (a) Example of a mask, and (b) applies multiple masks.}
    \label{fig:potentialscore}
\end{figure}

To determine the potential field, we computed the sum of several masks.
A typical mask is shown in Fig. \ref{fig:potentialscore}a, which considers the ball position as a point light source and excludes shadows cast by obstacles.
The passing location is determined flexibly by computing the sum of masks with other simple gradients, as shown in Fig. \ref{fig:potentialscore}b.

\subsection{Acceleration/deceleration control}
\begin{figure}[ttbp]
    \centering
    \includegraphics[width=0.9\linewidth]{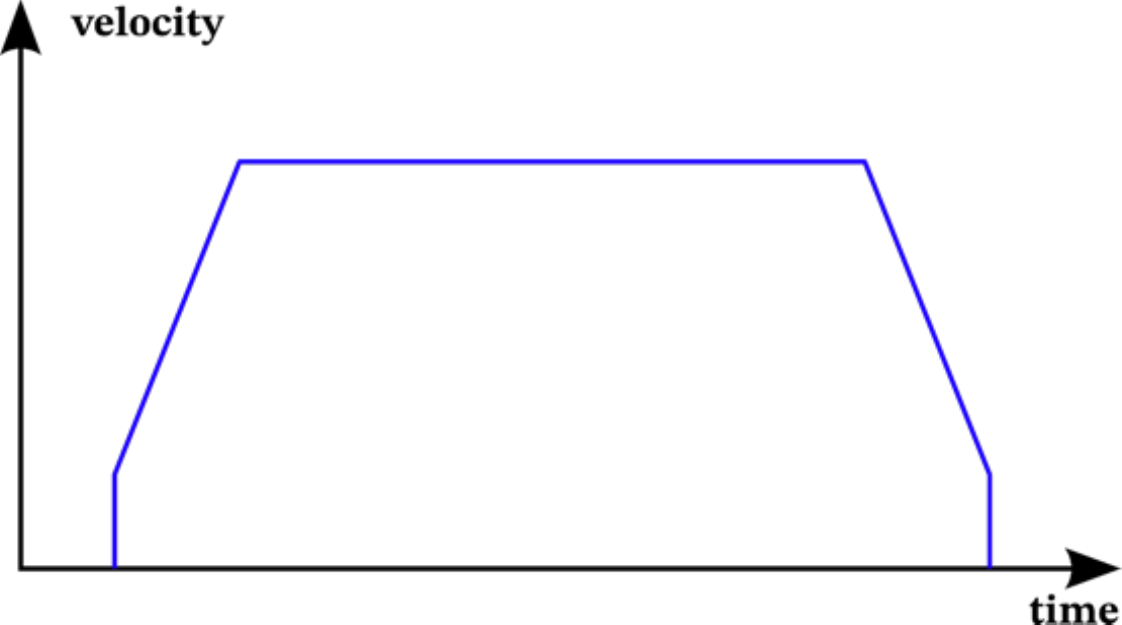}
    \caption{Trapezoidal acceleration/deceleration with both ends cut.}
    \label{fig:acc}
\end{figure}

Speed control is based on the trapezoidal acceleration/deceleration.
In the simple trapezoidal acceleration/deceleration, the speed is low in the low-speed range and short-distance movement.
For this reason, the trapezoidal acceleration/deceleration was cut at both ends, as shown in Fig. \ref{fig:acc}.

\subsection{Framework}
\begin{figure}[ttbp]
    \centering
    \includegraphics[width=0.9\linewidth]{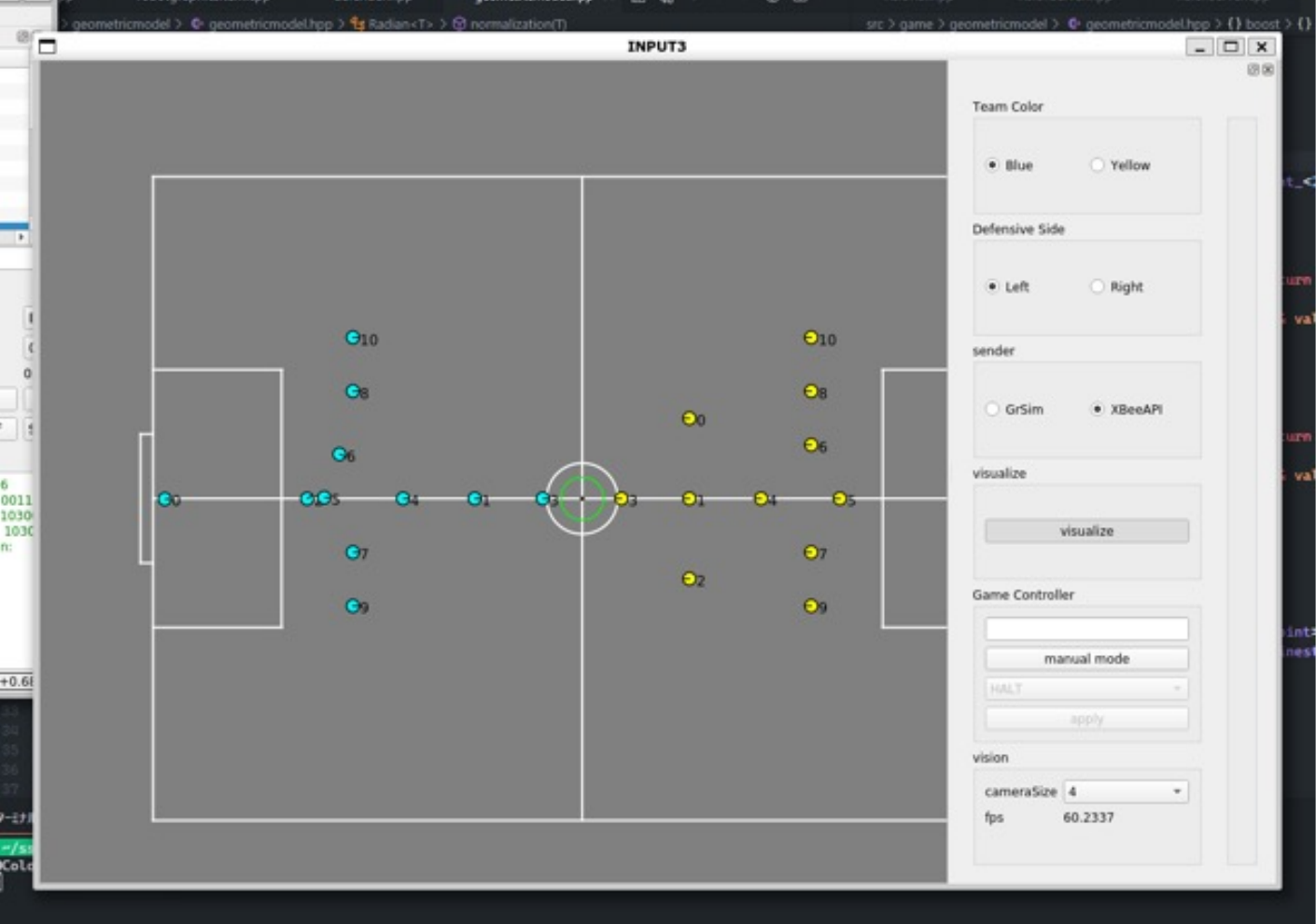}
    \caption{Software user interface.}
    \label{fig:ui}
\end{figure}

Qt was used as the software framework. The user interface developed in Qt is shown in Fig. \ref{fig:ui}.
Qt is used not only for the user interface, but also for UDP communication, serial communication for XBee, and the core "signal, slot.”

\subsection{Development Environment}
The software development was managed by GitLab.
We used the GitLab CI to automatically verify software builds and tests.
In addition, a system to output code documentation was configured using Doxygen.

\section{Introduction of robot development in cooperation with companies in the Nagaoka area}
\begin{figure}[ttbp]
    \centering
    \includegraphics[width=0.6\linewidth]{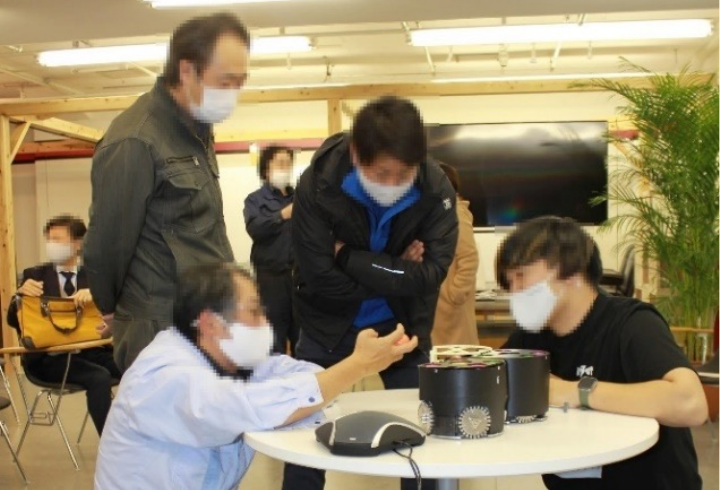}
    \caption{Discussions with partner companies.}
    \label{fig:discussion}
\end{figure}

Our team of students from Nagaoka is building a robot in collaboration with a company in Nagaoka.
Fig. \ref{fig:discussion} shows a discussion between students and a company in Nagaoka. The new Omni-Wheel, described in Section 4.1, was developed through this collaboration.

\begin{figure}[ttbp]
    \centering
    \includegraphics[width=0.6\linewidth]{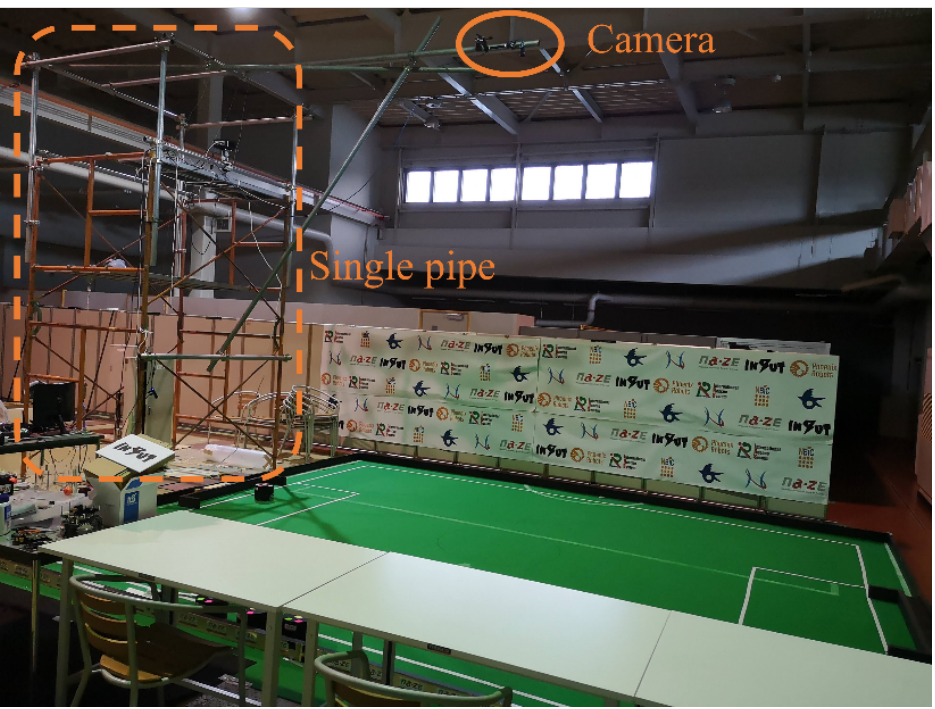}
    \caption{Camera mounted using single pipe.}
    \label{fig:camera}
\end{figure}

We also received support from a company in the Nagaoka area for the installation of cameras in the practice field of our team.
In general, SSL cameras are installed in a high position so that they can show a wide view of the field, but it is difficult to install a camera in a high position.
Therefore, we set up a scaffold made of a single pipe right next to the field and set up the camera at a height of approximately 4 m as shown in Fig. \ref{fig:camera}.

\section{Acknowledgement}
This robot was developed with the generous support and cooperation of NAZE and its member companies, who provided funding, design guidance, parts production, and field installation.
We would like to express our gratitude to Sanshin Co., NSS Co., and Ogawa Conveyor Corporation for their generous support.
We would like to take this opportunity to express our deepest gratitude.

%
%
%
%

\end{document}